\documentclass[twocolumn]{article}

\usepackage[usenames]{color} %used for font color
\usepackage{amssymb} %maths
\usepackage{amsmath} %maths
\usepackage[utf8]{inputenc} %useful to type directly diacritic characters
\usepackage{enumitem}
\usepackage{graphicx}
\usepackage[british]{babel}
\usepackage{csquotes}
\usepackage[backend=biber, style=apa]{biblatex}
\DeclareLanguageMapping{british}{british-apa}
\begin{document}

\title{Batch Selection for Parallelisation of Bayesian Quadrature}

\author{ Ed Wagstaff \\
University of Oxford
\and Saad Hamid \\
University of Oxford
\and  Michael Osborne \\
University of Oxford
}

\maketitle

\begin{abstract}
  Integration over non-negative integrands is a central problem in machine learning (e.g. for model averaging, (hyper-)parameter marginalisation, and computing posterior predictive distributions). Bayesian Quadrature is a probabilistic numerical integration technique that performs promisingly when compared to traditional Markov Chain Monte Carlo methods. However, in contrast to easily-parallelised MCMC methods, Bayesian Quadrature methods have, thus far, been essentially serial in nature, selecting a single point to sample at each step of the algorithm. We deliver methods to select batches of points at each step, based upon those recently presented in the Batch Bayesian Optimisation literature. Such parallelisation significantly reduces computation time, especially when the integrand is expensive to sample.
\end{abstract}

\section{Introduction}
Both within and without machine learning, we encounter integrals of the form
\begin{equation} \label{ZDef}
	Z = \int \ell(\boldsymbol{\theta})\pi(\boldsymbol{\theta})\text{d}\boldsymbol{\theta},
\end{equation}
where both $\ell(\boldsymbol{\theta})$ (e.g. a likelihood), and $\pi(\boldsymbol{\theta})$ (e.g a prior) are non-negative. These integrals are usually intractable, and we must therefore turn to approximation to solve them.

One possibility is the Laplace approximation. For multimodal distributions it can be repeated on the residuals -- the difference between the approximation and the true distribution -- to obtain a better approximation \parencite{Bornkamp2011}. However, this requires optimisation and is not invariant to changes of representation. Another possibility is variational inference which provides a lower bound for the value of the integral, but there is limited underpinning theory and optimisation of the lower bound can be difficult \parencite{Blei2018}. Markov Chain Monte Carlo (MCMC) methods have also seen broad use and success, but they are subject to a number of objections \parencite{OHagan1987}, the most pertinent of which is that they ignore the locations of samples from the likelihood when computing the integral. They therefore suffer from low sample efficiency and a poor $\frac{1}{\sqrt{N}}$ convergence rate. All of these approximation methods are unsuitable for applications where evaluating the desired likelihood is expensive (e.g. if we must make an estimate based on only a few dozen evaluations).

Bayesian Quadrature (BQ) offers an alternative framework for approaching the same integral problem. It belongs to a family of methods termed \emph{probabilistic numerics} \parencite{Hennig2015}, which view the approximation of intractable quantities as a problem of inference from limited data, to which the tools of probability theory can be applied. In BQ, a probabilistic model of the integrand (typically a Gaussian Process) is maintained, from which the integral is inferred. Gunter et al. \parencite*{Gunter2014} showed that BQ can achieve faster convergence -- both in wall-clock time and sample complexity -- than state of the art MCMC methods. Some deterministic quadrature techniques can be reinterpreted in terms of Bayesian Quadrature -- for instance, the trapezoid rule (with $\frac{1}{N}$ convergence) arises from BQ with a linear spline kernel \parencite{Karvonen2017}.

One drawback of BQ as developed thus far is that is not easily parallelised, in contrast to the comparatively straightforward situation with MCMC methods where, for instance, several independent Markov chains may be run in parallel. While the probabilistic model allows us to explore the space more efficiently, determining where next to evaluate the integrand depends on the current state of the model, i.e. on all evaluations made so far. Nonetheless, parallelisation is essential to realising competitive performance using modern computational hardware.

Using ideas applied in Batch Bayesian Optimisation, particularly those by Gonzalez et al. \parencite*{Gonzalez2015}, we propose a novel technique for parallelising BQ by selecting new points in batches rather than one-by-one.

\section{Bayesian Quadrature}

In Bayesian Quadrature, we work with a probabilistic model of an integrand which induces a probability distribution over the value of the integral. The early work on BQ \parencite{Diaconis1988, OHagan1991, Minka2000, Rasmussen2003} identified that Gaussian Processes offer a well-studied method for placing a probability distribution over functions, so we work with these to provide our probabilistic model.

Recall from Equation \eqref{ZDef} that we are interested in the quantity $\int \ell(\boldsymbol{\theta})\pi(\boldsymbol{\theta})\text{d}\boldsymbol{\theta}$. We typically choose to place our Gaussian process prior on the likelihood $\ell$, rather than on the product $\ell\cdot\pi$ (though we could choose to do either).

Given a Gaussian prior $\pi$ and a GP model for our likelihood,

\begin{subequations}
\begin{align}
\ell \mid \mathbf{y} &\sim \mathcal{GP}(m_\mathbf{y}, C_\mathbf{y})\\
m_\mathbf{y}(x) &= k(x, X)k(X, X)^{-1}\mathbf{y}\\
C_\mathbf{y}(x, x^\prime) &= k(x, x^\prime) - k(x, X)k(X, X)^{-1}k(X, x^\prime)
\end{align}
\end{subequations}

the value of the integral $Z = \int \ell(x)\pi(x)\text{d}x$ is Gaussian distributed -- since Gaussian Processes are closed under linear operators (see e.g. \cite{Rasmussen2006}), the GP on the function induces a normal distribution on the value of the integral $Z$. By linearity of expectation and bilinearity of covariance, the mean and variance of $Z$ are simply \parencite{Briol2015}:

\begin{subequations}
\begin{align} \label{eq:11}
\mathbb{E}\left( Z \mid \mathbf{y} \right) &= \int m_\mathbf{y}\left(x\right)\pi(x)\text{d}x\\
 \label{eq:12}
\text{var}\left(Z \mid \mathbf{y} \right) &= \iint C_\mathbf{y}\left(x, x^\prime\right)\pi(x)\pi(x^\prime) \text{d}x\text{d}x^\prime
\end{align}
\end{subequations}

These integrals are, for certain choices of prior, $\pi$, and kernel, $K$, analytic. In particular, this is true when $\pi$ is a Gaussian and $K$ is the squared exponential covariance, and we use this combination throughout this paper.

At a high level, the advantages of Bayesian Quadrature are two-fold. First, with the stronger prior of BQ, evaluations provide stronger constraints on the integrand, and therefore more information about the integral. Secondly, we can use the uncertainty of the GP model to inform our choice of where next to evaluate the integrand, thus exploring the space more efficiently. We do this by defining and maximising an \emph{acquisition function} which is high-valued at points which, according to our model, will reduce the uncertainty in the integral. In the standard BQ model, this second advantage does not apply -- note that the posterior covariance only depends on the \emph{locations} sampled, not on the function value at those points \parencite{Osborne2012}. However, as we shall see below, it is possible to develop more complex models where, among other advantages, the covariance of better-informed posteriors does depend on the sampled function values.

\subsection{Warped Bayesian Quadrature}

The Gaussian process framework has some restrictions which interfere with our ability to model our true beliefs about the functions we wish to work with. Standard Bayesian Quadrature does not allow us to express our belief that the function to be modelled (in our case a likelihood) is non-negative, since the distribution at any point is a Gaussian. Recent work has investigated applying a warping to the output space of the GP to address this \parencite{Osborne2012, Gunter2014, Chai2018}. Rather than modelling the likelihood $\ell$ itself as a GP, we model a transformed likelihood $g(\ell)$ such that the true likelihood $g^{-1}(g(\ell))$ is non-negative (i.e. such that the range of $g^{-1}$ is the positive reals). For example, $g=\text{log}(x)$ and $g=\sqrt{x}$ have been investigated. Though the final posterior on $\ell$ is not a GP, it is possible to derive a GP which closely approximates it.

For the purposes of this investigation we follow the method of Gunter et al. \parencite*{Gunter2014}, dubbed WSABI, which makes use of the transformation $g=\sqrt{x}$. In particular we use the approximation dubbed WSABI-L, which has the virtue of providing a closed-form estimate for the final integral rather than requiring Monte Carlo estimation. In WSABI, the acquisition function is simply the variance of the posterior model -- due to the form of the model, this automatically balances exploring areas of high uncertainty and exploiting in areas where the objective function is believed to be high-valued. WSABI has so far been considered only as a sequential process, selecting and evaluating one point at a time. We propose its  parallelisation by turning to batch methods.

\section{Batch methods}

Bayesian Quadrature as described above is an essentially serial process. That is, we must sample a single point at each step, before updating our model (and therefore our acquisition function) accordingly in order to recommend the next point. Since we expect each sample to be expensive, this is a serious bottleneck -- we would rather be able to work in parallel, taking several samples at once in order to gain the maximum amount of information about our integrand. In order to enable this, we turn to ideas which have been successfully used in Bayesian Optimisation.

The field of Bayesian Optimisation has a clear relationship to Bayesian Quadrature. In particular, both fields use a probabilistic model to guide exploration of a search space by defining an acquisition function to be maximised. In Bayesian Optimisation, so-called batch methods have been proposed as an effective way of allowing parallelisation of this exploration (see e.g. Ginsbourger et al. \parencite*{Ginsbourger2010}, Gonzalez et al. \parencite*{Gonzalez2015}, Nguyen et al. \parencite*{Nguyen2017}).

In batch methods, rather than selecting one point to sample at each step, we select a batch of points, usually of a fixed size $n$. Naively one could try to select the $n$ highest points of the acquisition function, but this has the obvious issue that these will all be essentially equal to the maximum of the acquisition function, leading to an uninformative batch -- the quality of the batch as a whole is more complex than simply the sum of the quality of each point individually. One way of viewing this issue is that we require an informative batch, and therefore we require a diverse batch.

To find the optimal batch we could, in principle, define an acquisition function over all sets of $n$ points (e.g. by considering the expected reduction in variance of our integral estimate) and maximise it, but in practice this is too expensive -- it increases the dimensionality of the problem by a factor of $n$ and the acquisition function is also likely to be more costly to compute. It is therefore common to turn to heuristics where each point is selected sequentially, with the acquisition function modified after each point is selected.

A simple method for performing this sequential selection is the so-called Kriging Believer strategy \parencite{Ginsbourger2010}. We ``sample'' our chosen point by setting its value to the posterior mean, update our Gaussian process model to take the new evidence into account, and select the most informative point from the posterior. We repeat this process until we have a sufficiently large batch, and then make the (real) evaluations at selected points and input them as observations into our model. This naturally forces further exploration of the space, as the region immediately around each ``sampled'' point will no longer be considered informative.

We may also consider a \emph{local penalisation} strategy which directly penalises the acquisition function around the chosen point (in Bayesian Optimisation, this strategy was introduced by Gonzalez et al. \parencite*{Gonzalez2015}). With this strategy, we don't use our full model, as in Kriging Believer, but instead directly modify the acquisition function to lower its value around each selected point, encouraging diversity in the final batch.

In the next section, we will consider the application of batch methods to Bayesian Quadrature.

\section{Method}

In this section we introduce, to our knowledge, the first Batch Bayesian Quadrature routine. We also introduce a novel version of local penalisation.

The core algorithm is simple -- we repeatedly select a batch of points, evaluate the integrand at these points, and input the returned data into our model. At each iteration, we can calculate an estimate of the integral using BQ. There is a question here of the stopping criterion. Theoretically, Bayesian Quadrature offers a principled stopping criterion since it returns a variance for its estimate of the integral, and we can simply evaluate the variance at each iteration and stop once it has reached a certain threshold. For the specific case of WSABI-L, the model is overconfident away from the data, and we expect that the BQ variance will therefore be erroneously low and unsuitable as a stopping criterion. This is a question that clearly merits further investigation, but for the purposes of this paper we simply set a fixed computational budget.

The most important step of the algorithm is the batch selection process. Above we described the Kriging Believer approach, and in general terms described local penalisation. We now outline our approach to local penalisation.

\subsection{Local Penalisation}

With local penalisation methods, there is a balance to be struck in how wide the penalisers are -- if our penalisation affects a broad region, we may erroneously exclude genuinely informative points. If, however, our penalisation affects a narrow region, we may not sufficiently encourage diversity, and produce a batch of points which are clustered too closely together. Gonzalez et al. \parencite*{Gonzalez2015} propose a multiplicative penaliser with a minimum at the sampled point. To determine how wide the penaliser should be, they make use of a Lipschitz constant, a measure of the maximum absolute slope of the acquisition function. In this method the Lipschitz constant is incorporated into a probablility density function which is used as part of the penaliser. The authors' proposition is less suited to quadrature than to optimisation, as it is not sufficiently exploratory -- it heavily penalises regions surrounding a selection which has a high variance. Thorough exploration is more important in quadrature, since the result depends on the global behaviour of the function rather than a single optimum. We therefore consider an alternative approach, attempting to place an upper bound on the true acquisition function (i.e. the acquisition function had the integrand been evaluated at the previous selections).

Recall that a function $f$ is said to be Lipschitz continuous if there is a constant $L$ such that, for all $x$, $y$,

\begin{equation}
||f(x) - f(y)|| \leq L ||x - y||.
\end{equation}

In our method, we apply the Lipschitz constant more directly. We expect our true acquisition function to be zero at the sampled point and increase away from it, and wish for our penalised function to reflect this. We find a local Lipschitz constant for a given point by locally optimising the norm of the gradient of the acquisition function. Viewing this as a bound on the slope of the acquisition function around that point, we then define a \emph{Lipschitz Cone} which is zero at the point and increases linearly away from the point. i.e. for a Lipschitz constant $L$ and point $x_0$, we define the Lipschitz Cone,

\begin{equation}
C(x) = L||x-x_0||.
\end{equation}

Our acquisition function is then taken to be the minimum of the original acquisition function and the Lipschitz Cones of all points so far added to the batch.

This minimum, however, is not differentiable everywhere. In particular, it is not differentiable where the cone intersects the original acquisition function (or where two cones intersect). These intersections may in fact be maxima of the penalised acquisition function, so ideally the penalised function would have stationary points on these intersection surfaces. We address this by applying a soft minimisation, rather than the true minimum, which leaves us with an acquisition function which is differentiable everywhere except at the centres of the Lipschitz cones (i.e. the already selected batch points).

\begin{figure}%[h!]
%\begin{center}
%\includegraphics[width=\linewidth,]{Penalisation.png}
\centering
\includegraphics[height=0.87\textheight]{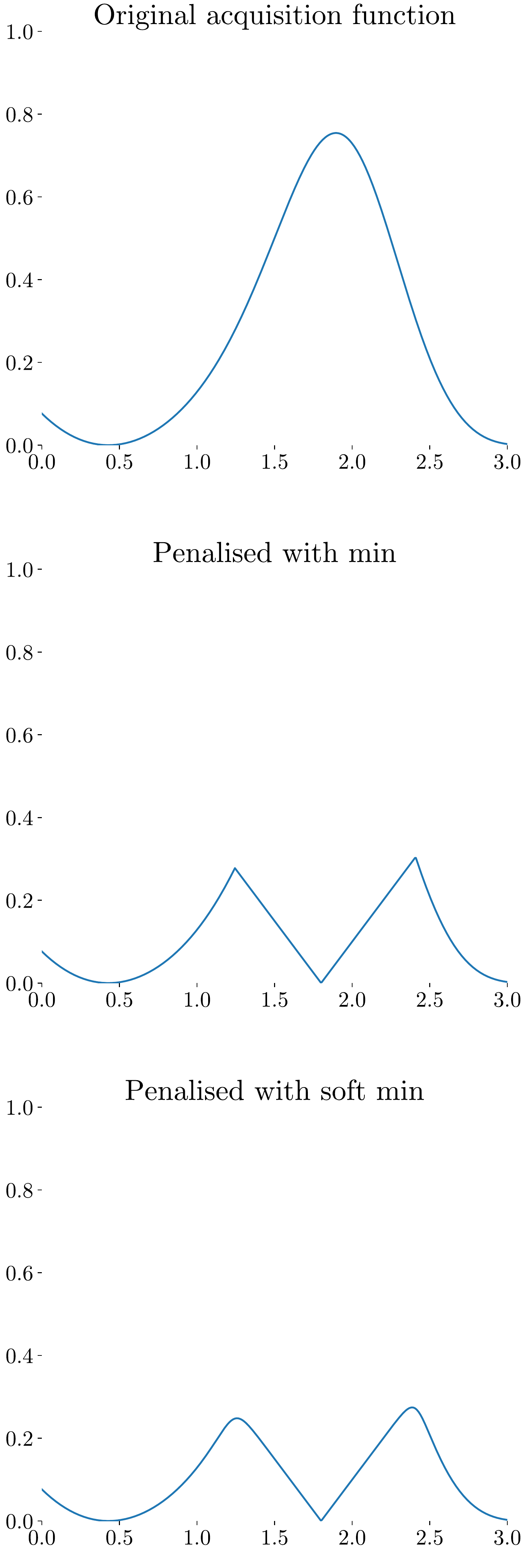}
%\end{center}
\caption{The penalisation of the acquisition function, showing the non-differentiable minimum, and the differentiable soft minimisation achieved by applying the $\ell_p$-norm for $p=-6$. Note that the maxima after penalisation are stationary points with the soft minimisation, but not with the true minimisation.}
\end{figure}

%\clearpage

Recall that the $\ell_p$ norm $||\cdot||_p$ is defined by

\begin{equation}
||x||_p = \left(\sum_i x_i^p\right)^\frac{1}{p}.
\end{equation}

It is well known that the $\ell_p$-norm for large $p$ approaches the $\ell_\infty$ norm, $||x||_\infty = \text{max}_i(x_i)$. Large values of $p$ can therefore be viewed as applying a kind of soft maximisation. Large negative values of $p$, on the other hand, apply a soft minimisation to the input vector.\footnote{For negative $p$ this is no longer a norm, but we refer to it as an $\ell_p$-norm for brevity.} We therefore take our final acquisition function to be an $\ell_p$-norm of the vector whose components are the function evaluation for each of the penalisers and the original acquisition function. There is a balance to be struck in the choice of $p$ here -- large $p$ will more closely resemble the true minimum of all the functions, but will also be more prone to numerical errors (these errors will appear particularly around the important regions where the true minimum is not differentiable). We found that small values of $p$ already came reasonably close to the true minimum function, and selected $p=-6$ for our experiments.

Since we expect the true acquisition function to be smooth, rather than resembling a sharp cone near evaluated points, we set the gradient of our Lipschitz cone to be a fraction of the Lipschitz constant -- this has the effect of encouraging (or rather not discouraging) exploration. We found $\frac{1}{2}$ to work well in practice. In principle, this parameter could be tuned to help determine the degree to which the algorithm explores diverse regions of the space, with smaller values encouraging exploration more strongly, but we did not investigate this possibility as part of this work.

\subsection{Optimising the acquisition function}

To optimise the acquisition function, we run a local optimiser multiple times from randomly selected starting locations. We sampled these locations according to the prior $\pi$ against which we are integrating. We chose the number of starting locations to be 10 times the dimensionality of the integrand.

\subsubsection{Implementing Multi-start Optimisation}

Our code for this method was written in python, using numpy and scipy. The most straightforward way to implement multi-start optimisation is to simply loop over the starting points and perform the optimisation many times. It is however possible to speed up this process when using numpy by taking advantage of fast vectorised computation in numpy rather than using slow python loops. The acquisition function can be evaluated at many points by passing these points as a numpy array, and this is many times faster than passing the points one at a time in a python loop. To take advantage of this in the optimisation process, we posed our multi-start optimisation problem as a single-start optimisation problem in higher dimensions. Each point in our high-dimensional space is the concatenation of multiple points in the original low-dimensional space. Denoting concatenation by $|$, we define a new function $g(x_1 | ... | x_n) = f(x_1) + ... + f(x_n)$. The Jacobian of this function is then obtained by concatenating the Jacobian of $f$ at each $x_i$, and we eliminate the python loop over starting points by simply starting this high-dimensional optimisation at $x_1 | ... | x_n$.

\section{Experiments}

\begin{figure}[b!]
%\begin{center}
%\includegraphics[width=\linewidth,]{InModel.png}
\centering
\includegraphics[width=\columnwidth]{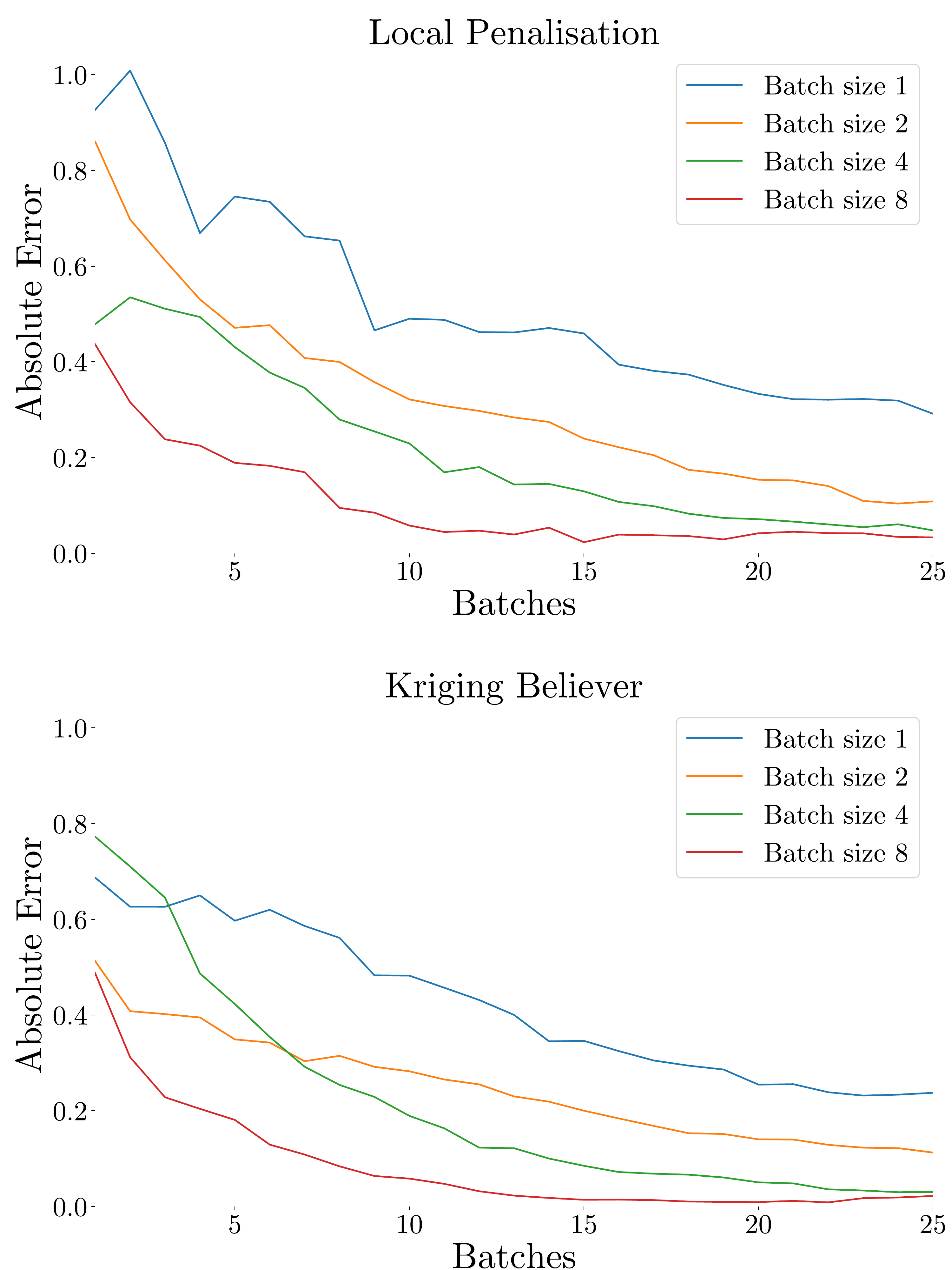}
%\end{center}
\caption{Convergence on an in-model integrand for different batch sizes and batch selection methods.}
\label{fig:inmodel}
\end{figure}

Prior work \parencite{Osborne2012, Gunter2014, Chai2018} has demonstrated that in terms of wall-clock time, BQ methods are competitive with MCMC. Here we investigate the degree to which adding batch methods can improve the performance of BQ in terms of the number of batches required (which, with an objective function which can be effectively parallelised,  is a better indicator of actual time taken than the number of samples taken). For a batch of size $n$, the best we can reasonably hope for is a factor-$n$ decrease in batch complexity against the sequential (i.e. batch size 1) method. We test both of our proposed batch selection methods on synthetic problems to assess the improvement achieved by batching, and compare both batch selection methods against MCMC methods in a test computing the model evidence of a Gaussian Process.

\begin{figure}[t!]
%\begin{center}
%\includegraphics[width=\linewidth,]{Synthetic.png}
\centering
\includegraphics[width=\columnwidth]{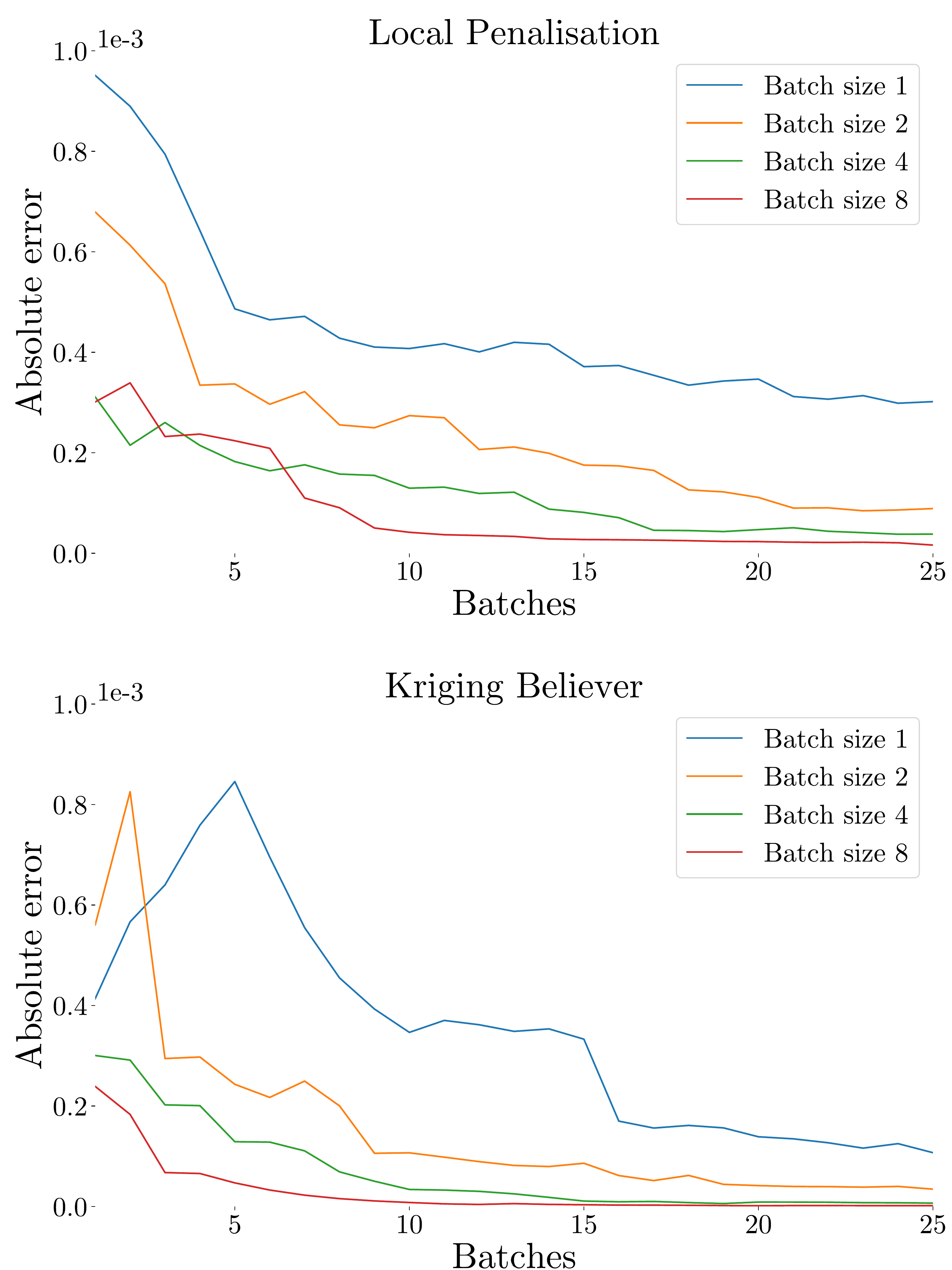}
%\end{center}
\caption{Convergence on a synthetic integrand for different batch sizes and batch selection methods.}
\label{fig:synthetic}
\end{figure}

\begin{figure}[hbt!]
%\begin{center}
%\includegraphics[width=\linewidth,]{Synthetic.png}
\centering
\includegraphics[width=0.98\columnwidth]{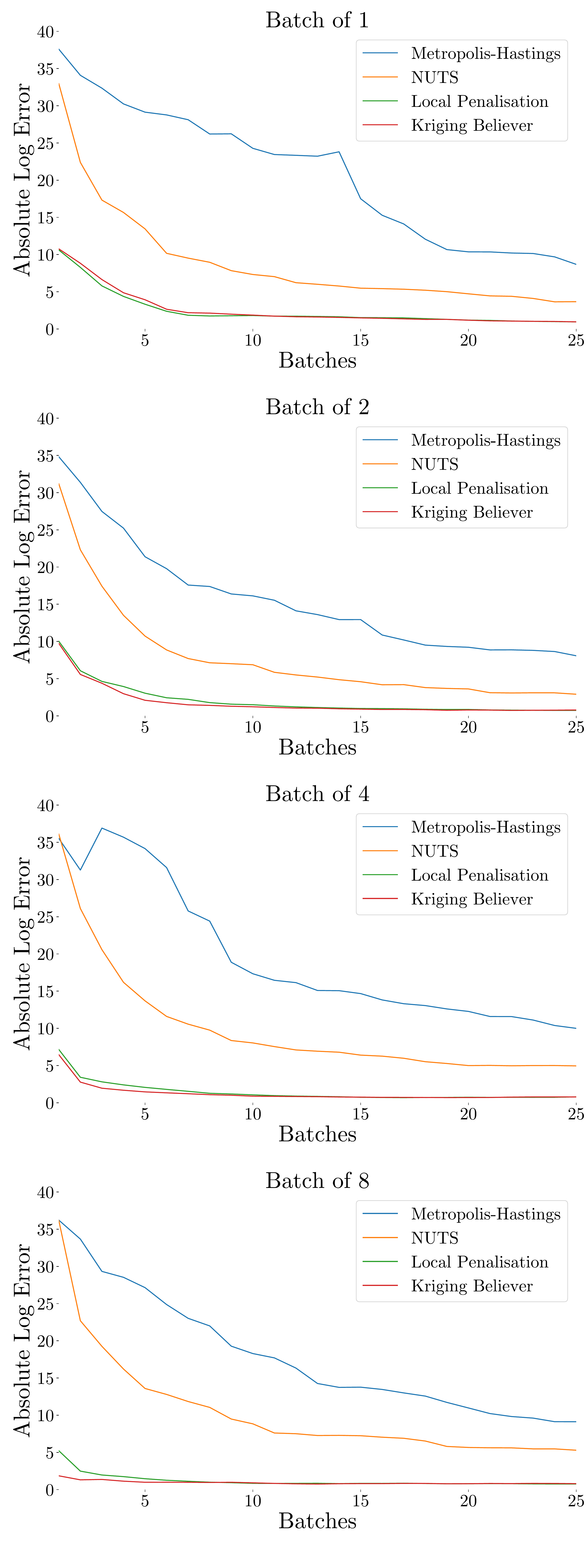}
%\end{center}
\caption{Computing the model evidence for a Gaussian Process.}\label{fig:model_evidence}
\end{figure}

Our first experiment is on an in-model integral, i.e. a sample from a Gaussian Process prior with the WSABI warping applied to it. We integrated this function with WSABI for a number of different batch sizes. Each sample was taken in 2 dimensions, and integrated against an isotropic Gaussian prior. Because we are working in 2 dimensions, the ground truth was obtained by simply evaluating the function on a fine grid and evaluating the integral naively. We plot the average absolute error over 10 runs of this experiment in Figure \ref{fig:inmodel}.

The in-model test demonstrates effective parallelisation of the process -- we can clearly see that larger batch sizes converge much more quickly to the true integral. We also see that both methods are similarly effective -- Kriging Believer and Local Penalisation show comparable performance.

Our second experiment is on a synthetic integral. Here we construct our integrand as a mixture of isotropic Gaussians.

\clearpage

We do this in 4 dimensions, varying the number of components between 10 and 15, setting their variance uniformly at random between 1 and 4, and setting their means uniformly at random within the box bounded by $[-3, 3]$ in all dimensions. We set the hyperparameters of the underlying GP in our model (which uses an isotropic squared exponential kernel) by optimising the marginal likelihood.

Again we test both of our batch selection methods, with a varying number of batch sizes. We plot the average absolute error over 10 runs of this experiment in Figure \ref{fig:synthetic}. The results are similar to the in-model test, demonstrating an effective speedup with increasing batch size, and again showing comparable performance between the two batch selection methods. 

Our final experiment is computing the model evidence for a Gaussian Process. The data for the GP was collected by running 20 iterations of Bayesian Optimisation on the Branin-Hoo function, a common benchmark function for Bayesian Optimisation \parencite{Jones2001}. For the optimisation we used a GP with an isotropic squared exponential covariance, giving us two hyperparameters to marginalise over -- we marginalise against a Gaussian prior on the hyperparameters. Since this integral is 2D, we once again obtained ground truth by evaluating on a fine grid.

In this experiment we also compare against two MCMC methods -- Metropolis-Hastings (Metropolis et al., \cite*{Metropolis1953}; \cite{Hastings1970}), a popular standard method, and NUTS \parencite{Hoffman2014}, a newer state-of-the-art method. Both of these methods were run with 500 samples of burn-in, and when comparing against our batched method with a batch size of $n$, we take the result obtained from running $n$ parallel chains. Figure \ref{fig:model_evidence} shows the average absolute error in the log of the model evidence, over 50 runs of this experiment. Once again it is clear that larger batch sizes converge more quickly for our method, and our method clearly outperforms the MCMC benchmarks.

All three experiments show improved performance as batch size increases, demonstrating that the selected batches are genuinely informative sets of points.

\section{Conclusion}

In this paper we have presented a routine for performing Bayesian Quadrature as a batched method, and introduced a novel form of local penalisation as part of the batch selection process. Our method allows parallelisation of methods which have so far only been available as serial processes. We have demonstrated that the performance of our method improves as the degree of parallelism increases, and shown that the method is competitive with state-of-the-art MCMC when the sampling budget is small.

\printbibliography
\end{document}